# New Edge Detection Technique based on the Shannon Entropy in Gray Level Images

Mohamed A. El-Sayed
Department of Mathematics,
Faculty of Science, Fayoum University, Egypt
mas06@fayoum.edu.eg

Tarek Abd-El Hafeez
Department of Computer Science,
Faculty of Science, Minia University, Egypt
Tarek.hemdan@Science.miniauniv.edu.eg

**Abstract:** Edge detection is an important field in image processing. Edges characterize object boundaries and are therefore useful for segmentation, registration, feature extraction, and identification of objects in a scene. In this paper, an approach utilizing an improvement of Baljit and Amar method which uses Shannon entropy other than the evaluation of derivates of the image in detecting edges in gray level images has been proposed. The proposed method can reduce the CPU time required for the edge detection process and the quality of the edge detector of the output images is robust. A standard test images, the real-world and synthetic images are used to compare the results of the proposed edge detector with the Baljit and Amar edge detector method. In order to validate the results, the run time of the proposed method and the pervious method are presented. It has been observed that the proposed edge detector works effectively for different gray scale digital images. The performance evaluation of the proposed technique in terms of the measured CPU time and the quality of edge detector method are presented. Experimental results demonstrate that the proposed method achieve better result than the relevant classic method.

## 1. INTRODUCTION

Edge detection is an important field in image processing. It can be used in many applications such as segmentation, registration, feature extraction, and identification of objects in a scene. Edge detection refers to the process of locating sharp discontinuities in an image. These discontinuities originate from different scene features such as discontinuities in depth, discontinuities in surface orientation, and changes in material properties and variations in scene illumination[1].

Thresholding becomes then a simple but effective tool to separate objects from the background. Examples of thresholding applications are document image analysis where the goal is to extract printed characters [2], [3], logos, graphical content, musical scores, map processing where lines, legends, characters are to be found [4], scene processing where a target is to detected [5], quality inspection of materials [6], [7]. Other applications include cell images [8], [9] and knowledge representation [10], segmentation of various image modalities for non-destructive testing (*NDT*) applications, such as ultrasonic images in [11], eddy current images [12], thermal images [13], X-ray computed tomography (*CAT*), laser scanning confocal microscopy [14], extraction of edge field [15], image segmentation in general [16], [17], spatio-temporal segmentation of video images [18] etc. The output of the thresholding operation is a binary image whose gray level of 0 (black) will indicate a pixel belonging to a print, legend, drawing, or target and a gray level of 1 (white) will indicate the background. Many operators have been introduced in the literature, for example Roberts, Sobel and Prewitt [19], [20], [21], [22], [23]. Edges are mostly detected using either the first derivatives, called gradient, or the second derivatives, called Laplacien. Laplacien is more sensitive to noise since it uses more information because of the nature of the second derivatives.

**Concept of entropy:** Entropy is a concept in information theory. Entropy is used to measure the amount of information. Entropy is defined in terms of the probabilistic behavior of a source of information. In accordance with this definition, a random event *A* that occurs with probability *P(A)* is said to contain

$$I(A) = \log[1/P(A)] = -\log[P(A)]$$

**Units of information:** The amount *I(A)* is called the self-information of event A. The amount of self-information of the event is inversely related to its probability. If *P(A)* = 1, then *I(A)* = 0 and no information is attributed to it. In this case, uncertainty associated with the event is zero. Thus, if the event always occurs, then no information would be transferred by communicating that the event has occurred. If *P(A)* = 0.8, then some information would be transferred by communicating that the event has occurred. The base of the logarithm determines the unit which is used to measure the information. If the base of the logarithm is 2, then unit of information is bit.

If *P(A)* = ½, then *I(A)* = -log$_2$(½) = 1 bit. That is, 1 bit is the amount of information conveyed when one of two possible equally likely events occurs. A simple example of such a situation is flipping a coin and communicating the result (Head or Tail). The basic concept of entropy in information theory has to do with how much randomness is in a signal or in a random event. An alternative way to look at this is to talk about how much information is carried by the signal. Entropy is a measure of randomness.

The histogram and the probability mass function (*pmf*) of the image are indicated, respectively, by *h(g)* and by *p(g)*, *g* = 0...*G*, where *G* is the maximum luminance value in the image, typically 255 if 8-bit quantization is assumed. If the gray value range is not explicitly indicated as [$g_{min}$, $g_{max}$] it will be assumed to extend from 0 to *G*. The cumulative probability function is defined as $P(g) = \sum_{i=0}^{g} p(i)$. It is assumed that the *pmf* is estimated from the histogram of the image by normalizing to the number of samples at every gray level. In the context of document processing, the foreground (object) is the set of pixels with luminance values less than T, while the background pixels have luminance value above this threshold. In *NDT* images the foreground area may consists of darker (more absorbent, denser etc.) regions or conversely of shinier regions, for example that hotter, more reflective, less dense etc. regions. In contexts where the object appears brighter than the background the definitions of the foreground and background will be simply toggled.

The foreground (object) and background *pmf*'s will be expressed as $p_f(g), 0 \leq g \leq T$, and $p_b(g), T+1 \leq g \leq G$, respectively, where *T* is the threshold value. The foreground and background area probabilities are calculated as:

$$P_f(T) = P_f = \sum_{g=0}^{T} p(g) , \quad P_b(T) = P_b = \sum_{g=T+1}^{G} p(g)$$

The Shannon entropy parametrically dependent upon the threshold value T for the foreground and background is formulated as:

$$H_f(T) = -\sum_{g=0}^{T} p_f(g) \log p_f(g), \quad H_b(T) = -\sum_{g=T+1}^{G} p_b(g) \log p_b(g)$$

The sum of these two is expressed as $H(T) = H_f(T) + H_b(T)$. When the entropy is calculated over the input image distribution *p(g)* (and not over the class distributions), then obviously it does not depend upon the threshold *T* and hence is expressed simply as *H*. For various other definitions of the entropy in the context of thresholding, with some abuse of notation, we will use the same symbols of $H_f(T)$ and $H_b(T)$ [24].

The proposed approach solves the problem of run time algorithm and the quality of the edge detector of the output images. In the proposed method, we have divided the original image into four regions and calculate the optimal threshold for each region, instead of using only one threshold like Baljit and Amar method [25]. A standard test images, the real-world and synthetic images are used to compare the results of the proposed edge detector with the Baljit and Amar edge detector operator. In order to validate the results, the run time of the proposed method and the pervious method are presented.

## 2. Threshold Value And Scheme For Edge Detection Using Baljit And Amar Method

Threshold value is used to transform a dataset containing values that vary over some range into a new dataset containing just two values. When a threshold value is applied on to the input data, then input values that fall below the threshold are replaced by one of the output values and input values that at or above the threshold are replaced by the other output value. Image thresholding [18] is a segmentation technique because it classifies pixels into two categories. Category1: Pixels whose gray level values fall below the threshold and category2: Pixels whose gray level values are equal or exceed the threshold. In gray level image, range of input dataset is [0,255]. After thresholding, output dataset contains only two values 0 and 255. Thus, thresholding creates a binary image. If $T$ is a threshold value, then any pixel $(x, y)$ for which $f(x, y) > T$ is called an object point, otherwise the pixel is called a background pixel. In general, the threshold can be chosen as the relation, $T=T[x, y, p(x, y), f(x, y)]$ where $f(x, y)$ is the gray level of the pixel $(x, y)$ and $p(x, y)$ denotes some local property of this pixel, for example, the average gray level of a neighborhood centered on $(x, y)$. A threshold image $h(x, y)$ is defined as $h(x, y) =1$ if $f(x, y)>T$, otherwise $h(x, y)=0$. Thus, pixels labeled 1 correspond to objects, whereas pixels labeled 0 correspond to the background. When $T$ depends only on $f(x, y)$ (only on gray level values), the threshold is called global. If $T$ depends on $f(x, y)$ and $p(x, y)$, the threshold is called local. If $T$ depends on the pixel position $(x, y)$ as well as $f(x, y)$ at that pixel position, then it is called dynamic or adaptive threshold. In proposed scheme to detect edges, global threshold value is used [25].

For edge detection, an image defined in the real world is considered to be a function of two real variables, for example, $f(x, y)$ with $f$ as the amplitude (brightness) of the image at the real coordinate position $(x, y)$. A spatial filter mask may be defined as a (template) matrix $w$ of size $m \times n$. Assume that $m = 2a+1$ and $n = 2b+1$, where $a, b$ are nonzero positive integers. Smallest meaningful size of the mask is $3 \times 3$.

• Classification of all pixels that satisfy the criterion of homogeneousness
• Detection of all pixels on the borders between different homogeneous areas

In this scheme, first create a binary image by choosing a suitable threshold value. Window is applied on the binary image. Set all window coefficients equal to 1 except centre, centre equal to $\times$.

Move the window on the whole binary image and find the probability of each central pixel of image under the window. Then, the entropy of each central pixel of image under the window is calculated as

$$H(centralPix\,el) = -p\,log(p)$$

Where, $p$ is the probability of central pixel of binary image under the window. Now, the probability of central pixel, $p = 4/9$ and the entropy of central pixel,

$$H(centralPix\,el) = -p\,log(p) = -(4/9)\,log(4/9) = 0.3604$$

If, for any other instance, the image under the window is: In this case, the probability of central pixel, $p = 2/9$ and the entropy of central pixel,

$$H(centralPix\,el) = -p\,log(p) = -(2/9)\,log(2/9) = 0.3342$$

When the probability of central pixel, $p=1$, then the entropy of this pixel is zero. If the gray level of all pixels under the window homogeneous, $p=1$ and $H=0$. In this case, the central pixel is not an edge pixel. Thus, the central pixel with entropy greater than and equal to 0.2441 is an edge pixel, otherwise not [25].

## 3. The Proposed Algorithm:

We study the Baljit and Amar [25] scheme for edge detector and threshold algorithms and record the iteration number for calculating the optimal threshold value through 44 images and record the iteration number for each value of the initial estimate for $T$. we found that the minimum number of iterations to obtain the $T$ is in range [80-140] as shown in Figure 1, where the $T$ values at the $X$-axis and the number of iterations in $Y$-axis.

This range will satisfy the best results of the optimal threshold, and ensure that the selected threshold will produce robust results. We use this range of initial t in the modified algorithm for edge detector instead of using all the range [0-255]. This gives small iteration numbers and less run time than the previous method. The modification also contains the technique of computation.
We replace many used matrices in Baljit and Amar method with only variable $T\_new$ that record the best threshold in order to obtain the optimal threshold at the end of the procedure. This modification will reduce the run time of computations.

Also, we divide the image into four equal regions; each region will have its optimal threshold. This means that palpability of the error at the worst case to obtain the edge detection will be decrease with 1/4 of the resultant image.

The results showed that, the average of individual iterations of each region is less than the average iterations of complete image as the previous work.

The following procedures summarize the proposed technique for calculating the optimal threshold and the edge detector algorithm:

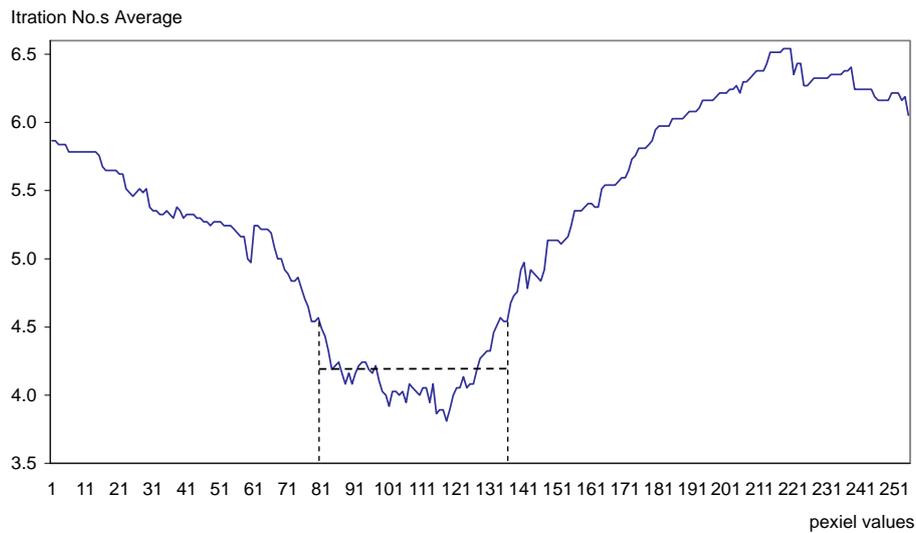

Figure 1. Minimum of iterations to obtain *T*.

*% The Threshold function*

function [f,T_rnd,T,ItrNo]=Thrshold_procedure(A);

[M N]=size(A);

T_rnd = randint(1,1,[80,140]);

T_new=T_rnd;

T_old=0;

ItrNo=0;

while (T_new ~= T_old)

  ItrNo = ItrNo + 1;

  Ta=0; Na=0; Tb=0; Nb=0;

  for i=1 : M

    for j=1 : N

      if (A(i,j) > T_new)    Ta = Ta + A(i,j);   Na = Na+1;    else    Tb = Tb + A(i,j);    end;

    end;

  end;

  Nb = M*N-Na;

  if (Na>0)&(Nb>0) Ta=floor(Ta/Na); Tb=floor(Tb/Nb); end;

  T_old=T_new;

  T_new=floor((Ta + Tb)/2);

end;

T=T_new;

f=zeros(M,N);

for i=1:M ;for j=1:N; if (A(i,j) >= T) f(i,j)=1;end; end; end;

*% The Edge Detector function*

```
function [g]=EdgeDetector_procedure(f);
[row column]=size(f);
g=f;
m = 3;  n = 3; a0 = (m-1)/2; b0 = (n-1)/2;
for y = b0+1 : column-b0;
   for x = a0+1 : row-a0;
      sum1 = 0;
      for k=-b0:b0;
         for j=-a0:a0;
            if ( f(x,y) == f (x+j,y+k) ) sum1=sum1+1; end;
         end;
      end;
      if ( sum1>6 ) g(x,y)=0; else g(x,y)=1; end;
   end;
end;
```

*% The  Proposed Algorithm*

```
clc;
clear all;
FileName1 = 'lena.tif';
disp([ FileName1 ]);
A = double (imread(FileName1));
[M N] = size(A);
Xc=double (int8(M/2));
Yc=double (int8(N/2));

% Calling Thrshold_procedure for calculate the threshold of 4 regions in A.
[f1,T_rnd1,T1,ItrNo1]=Thrshold_procedure(A( 1:Xc, 1:Yc));
[f2,T_rnd2,T2,ItrNo2]=Thrshold_procedure(A( 1:Xc, Yc+1:N));
[f3,T_rnd3,T3,ItrNo3]=Thrshold_procedure(A( Xc+1:M, 1:Yc));
[f4,T_rnd4,T4,ItrNo4]=Thrshold_procedure(A( Xc+1:M, Yc+1:N));

% Calling EdgeDetector_procedure for calculate the edge detector of f.
f= [f1 f2 ; f3 f4];
g= EdgeDetector_procedure(f);
% Output the image , results, Ti* and iteration numbers.
figure; imshow(g);
fprintf(' T1* =%3d  T2* =%3d  T3* =%3d  T4* =%3d , \n',T1,T2,T3,T4);
fprintf(' Itr1=%3d  Itr2=%3d  Itr3=%3d  Itr4=%3d,  \n',ItrNo1,ItrNo2,ItrNo3,ItrNo4);
disp([ 'finshed...!!!' ]);
```

**4. Experimental Results:**

We run the Baljit and Amar method and the proposed algorithm 10 times for each image and the different optimal values of the threshold are presented. Figure 2 shows sample of results of both algorithms. The first column shows the original images, the second column shows the histogram of each image, the third and the fourth columns show the worst and the best cases of the Baljit and Amar method algorithm, respectively. The fifth and the last columns show the worst and the best cases of the proposed algorithm respectively.

As shown in Figure 2, it has been observed that the proposed edge detector works effectively for different gray scale digital images and the performance of the proposed edge detection scheme is found to be satisfactory for all the test images as compare to the performance of Baljit and Amar algorithm.

| Original image | Hist | Baljit and Amar worst case | Baljit and Amar best case | Proposed worst case | Proposed best case |
|---|---|---|---|---|---|
| 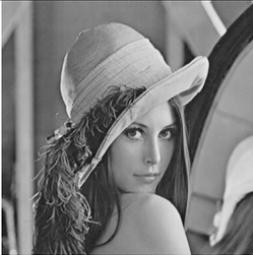 | 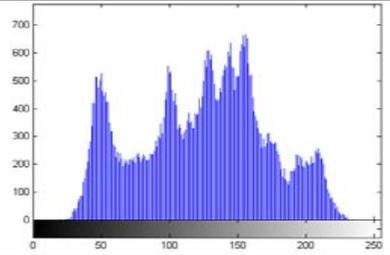 | 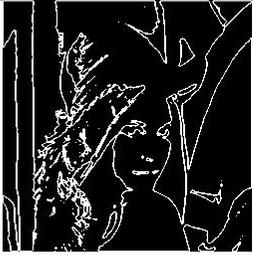 | 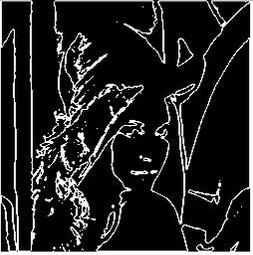 | 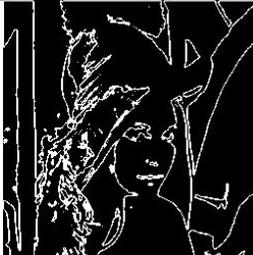 | 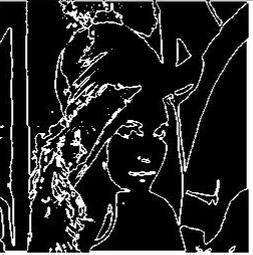 |
| Lena | | T=116 | T=117 | T=123,_113_103_120 | T130_113_104_120 |
| 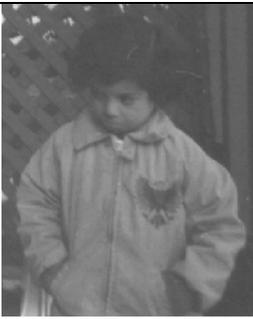 | 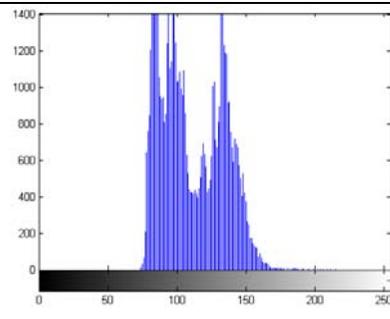 | 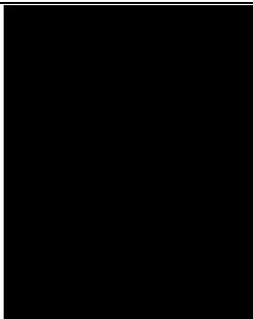 | 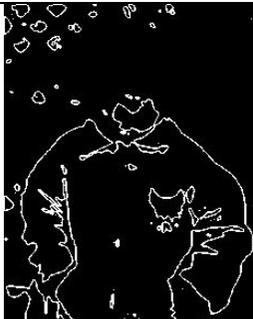 | 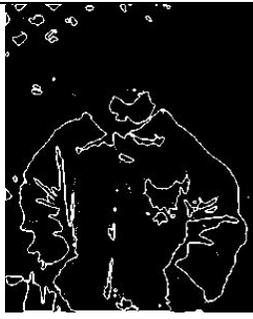 | 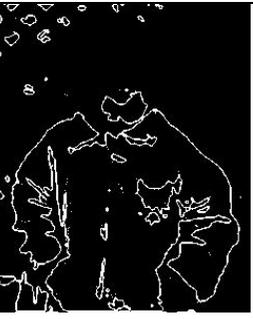 |
| Pout | | T=55 | T=113 | T=116,_111_118_113 | T116_111_119_114 |
| 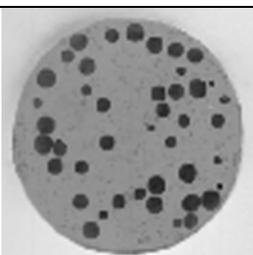 | 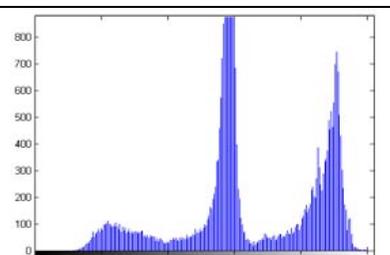 | 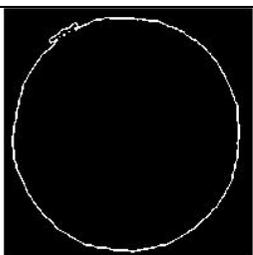 | 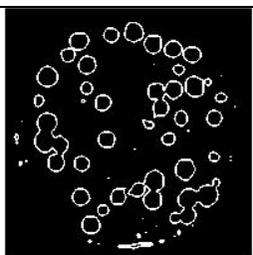 | 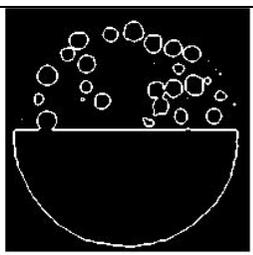 | 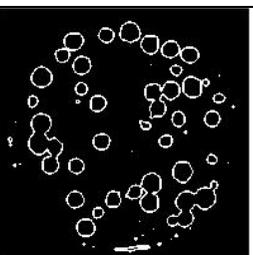 |
| Shot1 | | T=173 | T=125 | T=25,127,176_177 | T=125_126_123_124 |

Figure 2. Comparing of the performance of Baljit and Amar algorithm and the proposed algorithm for some gray scale digital images, Lena, Pout and Shot1 images .

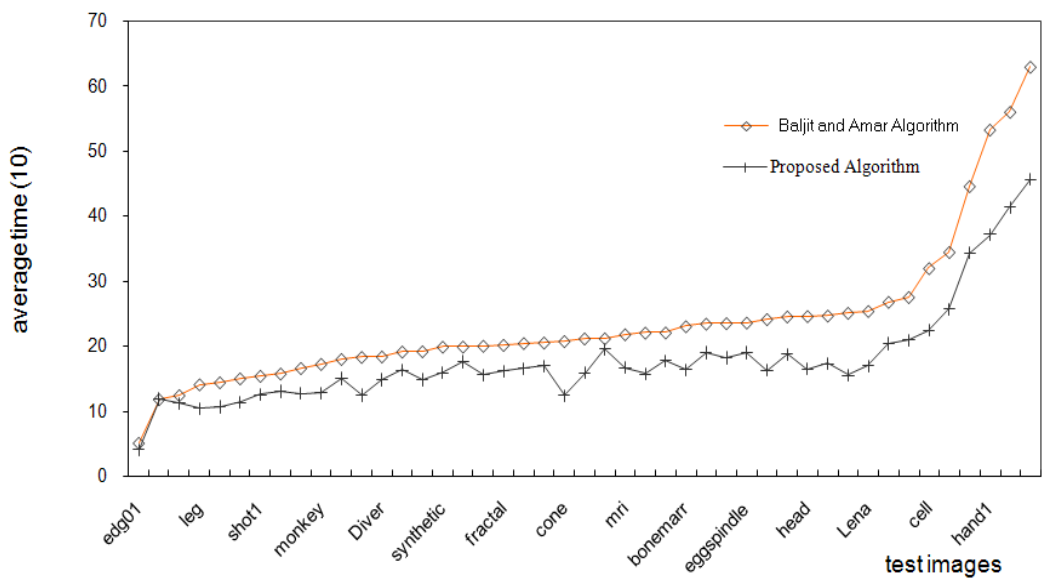

Figure 3 The cpu time of the proposed algorithm and Baljit and Amar algorithm for the test images .

We run the pervious and the proposed algorithm 10 times for each image and the average run time is calculated. Figure 3 shows that the proposed algorithm takes short time than the Baljit and Amar algorithm.

CONCLUSION

In this paper, an attempt is made to develop a new technique for edge detection. Experiment results have demonstrated that the proposed scheme for edge detection works satisfactorily for different gray level digital images. The proposed approach solves the problem of run time algorithm and the quality of the edge detector of the output images. In the proposed method, we have divided the original image into four regions and calculate the optimal threshold for each region, instead of using only one threshold like Baljit and Amar method. A standard test images are used to compare the results of the proposed edge detector with the Baljit and Amar edge detector operator. In order to validate the results, the run time of the proposed and the pervious method are presented. It has been observed that the proposed edge detector works effectively for different gray scale digital images. The results of this study were quite promising. The work is under further progress to examine the performance of the proposed edge detector for different gray level images affected with more than 4 regions of segmentations.